\documentclass[10pt,twocolumn,letterpaper]{article}

\usepackage{wacv}
\usepackage{times}
\usepackage{epsfig}
\usepackage{graphicx}
\usepackage{amsmath}
\usepackage{amssymb}
\usepackage{CJKutf8}

\def\wacvPaperID{312} 

\wacvfinalcopy 

\ifwacvfinal
\usepackage[breaklinks=true,bookmarks=false]{hyperref}
\else
\usepackage[pagebackref=true,breaklinks=true,colorlinks,bookmarks=false]{hyperref}
\fi

\begin{document}
\raggedbottom


\title{Transfer Learning for Pose Estimation of Illustrated Characters}

\author{
Shuhong Chen\\
University of Maryland - College Park\\
{\tt\small shuhong@cs.umd.edu}
\and
Matthias Zwicker\\
University of Maryland - College Park\\
{\tt\small zwicker@cs.umd.edu}
}

\maketitle
\thispagestyle{empty}

\begin{abstract}
Human pose information is a critical component in many downstream image processing tasks, such as activity recognition and motion tracking.  Likewise, a pose estimator for the illustrated character domain would provide a valuable prior for assistive content creation tasks, such as reference pose retrieval and automatic character animation.  But while modern data-driven techniques have substantially improved pose estimation performance on natural images, little work has been done for illustrations.  In our work, we bridge this domain gap by efficiently transfer-learning from both domain-specific and task-specific source models.  Additionally, we upgrade and expand an existing illustrated pose estimation dataset, and introduce two new datasets for classification and segmentation subtasks.  We then apply the resultant state-of-the-art character pose estimator to solve the novel task of pose-guided illustration retrieval.  All data, models, and code will be made publicly available.
\end{abstract}

\section{Introduction}

Human pose estimation is a foundational computer vision task with many real-world applications, such as activity recognition \cite{pesurvey}, 3D reconstruction \cite{arch}, motion tracking \cite{kinect}, virtual try-on \cite{tryon}, person re-identification \cite{reid}, etc.  The generic formulation is to find, in a given image containing people, the positions and orientations of body parts; typically, this means locating landmark and joint keypoints on 2D images, or regressing for bone transformations in 3D.

The usefulness of pose estimation is not limited to the natural image domain; in particular, we focus on the domain of illustrated characters.  As pose-guided motion retargeting of realistic humans rapidly advances \cite{dance}, there is growing potential for automatic pose-guided animation \cite{hamada}, a traditionally labor-intensive task for both 2D and 3D artists.  Pose information may also serve as a valuable prior in illustration colorization \cite{twostage}, keyframe interpolation \cite{animeinterp}, 3D character reconstruction \cite{buchanan} and rigging \cite{rignet}, etc.

With deep computer vision, we have been able to leverage large-scale datasets \cite{coco, mpipose, surreal} to train robust estimators of human pose \cite{rcnn, paf, alphapose}.  However, little work has been done to solve pose estimation for illustrated characters.  Previous pose estimation work on illustrations by Khungurn \etal \cite{manpu2016} presented a 2D keypoint detector, but relied on a publicly-unavailable synthetic dataset and an ImageNet-trained backbone.  In addition, the dataset they collected for supervision lacked variation, and was missing keypoints and bounding boxes required for evaluation under the more modern COCO standard \cite{coco}.

Facing these challenges, we constructed a 2D keypoint detector with state-of-the-art performance on illustrated characters, built upon domain-specific components and efficient transfer learning architectures.  We demonstrate the effectiveness of our methods by implementing a novel illustration retrieval system.  Summarizing, we contribute:
\begin{itemize}
    \item A state-of-the-art pose estimator for illustrated characters, transfer-learned from both domain-specific and task-specific source models.  Despite the absence of synthetic supervision, we outperform previous work by 10-20\% PDJ@20 \cite{manpu2016}.
    \item An application of our proposed pose estimator to solve the novel task of pose-guided character illustration retrieval.
    \item Datasets for our model and its components, including: an updated COCO-compliant version of Khungurn \etal's \cite{manpu2016} pose dataset with 2x the number of samples and more diverse poses; a novel 1062-class Danbooru \cite{danbooru2020} tagging rulebook; and a character segmentation dataset 20x larger than those currently available.
\end{itemize}

\begin{figure*}[!htb]
\begin{center}
\includegraphics[width=0.9\linewidth]{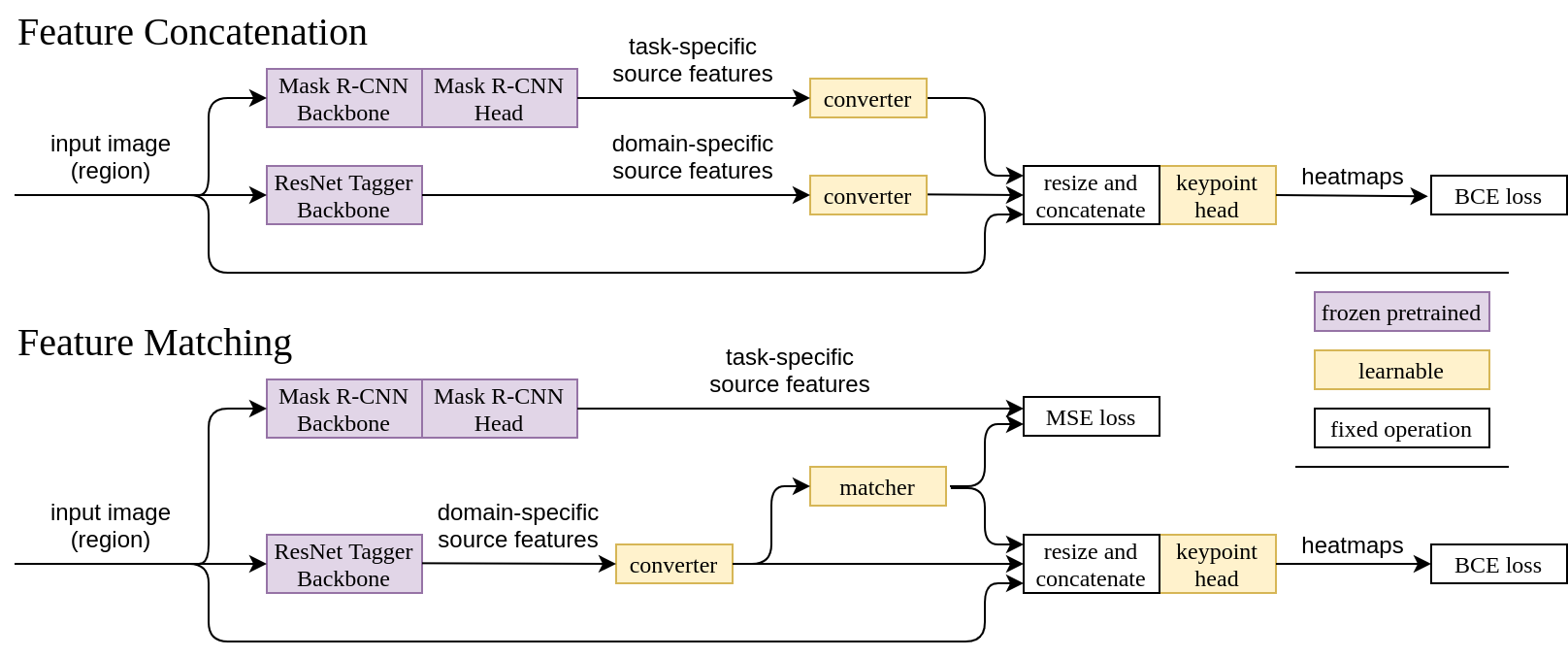}
\end{center}
\caption{A schematic outlining our two transfer learning architectures: feature concatenation, and feature matching.  Note that source feature specificity is with respect to the target; i.e. task-specific means ``related to pose estimation" and domain-specific means ``related to illustrations".  Feature converters and matchers are convolutional networks that learn to mimic or re-appropriate pretrained features, respectively.  While both designs require the pretrained Mask R-CNN components during training, feature matching discards them during inference, instead relying on the trained matcher network.  ``BCE" refers to binary cross-entropy loss.}
\label{fig:schematic}
\end{figure*}

\section{Related Work}

\subsection{The Illustration Domain}
Though there has been work on caricatures and cartoons \cite{caricature, puppet}, we focus on anime/manga-style drawings where characters tend to be less abstract.  While there is work for more traditional problems like lineart cleaning \cite{mastersketch} and sketch extraction \cite{mangastruct}, more recent studies include sketch colorization \cite{twostage}, illustration segmentation \cite{danbooregion}, painting relighting \cite{relight}, image-to-image translation with photos \cite{ugatit}, and keyframe interpolation \cite{animeinterp}.

Available models for illustrated tasks typically rely on small manually-collected datasets.  For example, the AniSeg \cite{aniseg} character segmenter is trained on less than $1{,}000$ examples.  While larger datasets are becoming available (e.g. Danbooru \cite{danbooru2020} now with 4.2m tagged illustrations), the labels are noisy and long-tailed, leading to poor model performance \cite{rf5, deepdanbooru}.  Works requiring pose information may use synthetic renders of anime-style 3D models \cite{manpu2016, hamada}, but the models are usually not publicly available.  In this work, we present a cleaner tag classification task, a large character segmentation dataset, and an upgraded COCO keypoint dataset; these will all be made available upon publication, and may serve as a valuable prior for other tasks.

\subsection{Transfer Learning \& Domain Adaptation}
Transfer learning and domain adaptation have been defined somewhat inconsistently throughout the vision and natural language processing literature \cite{dasurvey, daume}, though generally the former is considered broader than the latter.  In this paper, we use the terms interchangeably, referring to methods that leverage information from a number of related source domains and tasks, to a specific target domain and task.  Typically, much more data is available for the source than the target, motivating us to transfer useful related source knowledge in the absence of sufficient target data \cite{dasurvey}.  For deep networks, the simplest practice is to pretrain a model on source data, and fine-tune its parameters on target data; however, various techniques have been studied that work with different levels of target data availability.

Much of the transfer learning work in vision focuses on extreme cases with significantly limited target domain data, with emphasis around the task of image classification.  In the few-shot learning case, we may be given as few as ten (or even one) samples from the target, inviting methods that embed prototypical target data into a space learned through prior source knowledge \cite{fslsurvey}.  In particular, it is common to align parameters of feature extractors across domains, by directly minimizing pairwise feature distances or by adversarial domain discrimination \cite{luo2017label, adda}.  If the source and target are similar enough, it is possible to perform domain adaptation in the complete absence of labeled target data.  This can be achieved by matching statistical properties of extracted features \cite{coral}, or by converting inputs between domains through cycle-consistent image translation \cite{cycada}.

\subsection{Pose Estimation}
With the availability of large-scale human pose datasets \cite{coco, mpipose}, the vision community has recently been able to make great strides in pose estimation.  A naive baseline was demonstrated by Mask R-CNN \cite{rcnn}, which extended their detection and segmentation framework to predict single-pixel masks of joint locations.  Other work such as RMPE take an approach tailored to pose estimation, deploying spatial transformer networks with pose-guided NMS and region proposal \cite{alphapose}.  Around the same time, OpenPose proposed part affinity fields as a bottom-up alternative to the more common heatmap representation of joints \cite{paf}.  Human pose estimation work continues to make headway, extending beyond keypoint localization to include dense body part labels \cite{densepose} and 3D pose estimation \cite{human3m, up3d, lixel}.

\subsection{Pose Estimation Transfer}
Most transfer learning for pose estimation adapts from synthetically-rendered data to natural images.  For example, by using mocaps and 3D human models, SURREAL \cite{surreal} provides 6 million frames of synthetic video, complete with a variety of datatypes (2D/3D pose, RGB, depth, optical flow, body parts, etc.).  CNNs may be able to directly generalize pose from synthesized images \cite{surreal}, and can further close the domain gap using other priors like motion \cite{sim2real}.  Outside of synthetic-to-real, Cao \etal \cite{animals} explore domain adaptation for quadruped animal pose estimation, achieving generalization from human pose through adversarial domain discrimination with pseudo-label training.

The closest prior work to our topic was done by Khungurn \etal \cite{manpu2016}, who collected a modest AnimeDrawingsDataset (ADD) of 2k character illustrations with joint keypoints, and a larger synthetic dataset of 1 million frames rendered from MikuMikuDance (MMD) 3D models and mocaps.  Unfortunately, the MMD dataset is not publicly available, and ADD contains mostly standard forward-facing poses.  In addition, ADD is missing bounding boxes and several face keypoints, which are necessary for evaluation under the modern COCO standard \cite{coco}.  We remedy these issues by training a bounding box detector from our new character segmentation dataset, labeling missing annotations in ADD, and labeling 2k additional samples in more varied poses.

Khungurn \etal perform transfer from an ImageNet-pretrained GoogLeNet backbone \cite{googlenet} and synthetic MMD data.  In the absence of MMD, we instead transfer from a stronger backbone trained on a new illustration-specific classification task, as well as from a task-specific model pretrained on COCO keypoints.  We use our subtask models and data to implement a number of transfer techniques, from naive fine-tuning to adversarial domain discrimination.  In doing so, we significantly outperform Khungurn \etal on their reported metrics by 10-20\%.

\section{Method \& Architectures}
We provide motivation and architecture details for two variants of our proposed pose estimator (feature concatenation and feature matching), as well as two submodules critical for their success (a class-balanced tagger backbone and a character segmentation model).  Architectures for baseline comparison models are described in Sec. \ref{sec:archs}.

\subsection{Pose Estimation Transfer Model}
We present two versions of our final model: feature concatenation, and feature matching.   In this section, we assume that region proposals are given by a separate segmentation model (Sec. \ref{sec:bbox}), and that the domain-specific backbone is already available (Sec. \ref{sec:tagger}); here, we focus on combining source features to predict keypoints (Fig. \ref{fig:schematic}).

The goal is to perform transfer simultaneously from both a domain-specific classification backbone (Sec \ref{sec:tagger}) and a task-specific keypoint model (Mask R-CNN \cite{rcnn}).  Here, we chose Mask R-CNN as it showed significantly better out-of-the-box generalization to illustrations than OpenPose \cite{paf} (Tab. \ref{tab:inferperf}).  Taking into account that the task-specific model already achieves mediocre performance on the target domain, the feature concatenation model simply stacks features from both sources (Fig. \ref{fig:schematic}).  In order to perform the concatenation, it learns shallow feature converters for each source to decrease the feature channel count and allow bilinear sampling to a common higher resolution.  The combined features are fed to the head, consisting of a shallow converter and two ResNet blocks.

The final output is a stack of 25 heatmaps, 17 for COCO keypoints and 8 for auxiliary appendage midpoints (following Khungurn \etal \cite{manpu2016}).  We apply pixel-wise binary cross-entropy loss on each heatmap, targeting a normal distribution centered on the ground-truth keypoint location with standard deviation proportional to the keypoint's COCO OKS sigma \cite{coco}; the sigmas for auxiliary midpoints are averaged from endpoints of the body part.  At inference, we gaussian-smooth the heatmaps and take the maximum pixel value index as the keypoint prediction.

Although feature concatenation produces the best results (Tab. \ref{tab:inferperf}), it is very inefficient.  At inference, it must maintain the parameters of both source models, and run both forward models for each prediction; Mask R-CNN is particularly expensive in this regard.  We thus also provide a feature matching model, inspired by the methods used in Luo \etal \cite{luo2017label}.  As shown in Fig. \ref{fig:schematic}, we simultaneously train an additional matching network that predicts features from the expensive task-specific model using features from the domain-specific model.  Though matching may be optimized with self-supervision signals such as contrastive loss \cite{contrastive}, we found that feature-wise mean-squared error is suitable.  Given the matcher, the pretrained Mask R-CNN still helps training, but is not necessary at inference.  Despite its simplicity, feature matching retains most performance benefits from both source models, while also being significantly lighter and faster than the concatenation architecture.

\subsection{ResNet Tagger} \label{sec:tagger}
The domain-specific backbone for our model (Fig. \ref{fig:schematic}) is a pretrained ResNet50 \cite{ResNet} fine-tuned as an illustration tagger.  The tagging task is equivalent to multi-label classification, in this case predicting the labels applied to an image by the Danbooru imageboard moderators \cite{danbooru2020}.  The 392k unique tags cover topics including colors, clothing, interactions, composition, and even copyright metainfo.

Khungurn \etal \cite{manpu2016} use an ImageNet-trained GoogLeNet \cite{googlenet} backbone for their illustrated pose estimator, but we find that Danbooru fine-tuning significantly boosts transfer performance.  There are publicly-available Danbooru taggers \cite{rf5, deepdanbooru}, but both their classification performance and feature learning capabilities are hindered by uninformative target tags and severe class imbalance.  By alleviating these issues, we achieve significantly better transfer to pose estimation.

Most available Danbooru taggers \cite{rf5, deepdanbooru} take a coarse approach to defining classes, simply predicting the several thousand (6-7k) most frequent tags.  However, many of these tags represent contextual information not present in the image; e.g. neon\_genesis\_evangelion (name of a franchise), or alternate\_costume (fanmade/non-canon clothes).  We instead only allow tags explicitly describing the image (clothing, body parts, etc.).  Selecting tags by frequency also introduces tag redundancy and annotator disagreement.  There are many high-frequency tags that share similar concepts, but are annotated inconsistently; e.g. hand\_in\_hair, adjusting\_hair, and hair\_tucking have vague wiki definitions for taggers, and many color tags are subjective (aqua\_hair vs. blue\_hair).  To address these challenges, we survey Danbooru wikis to manually develop a rulebook of tag groups that defines more explicit and less redundant classes.

Danbooru tag frequencies form a long-tailed distribution, posing a severe class imbalance problem.  In addition to filtering out under-tagged images (detailed in Sec. \ref{sec:taggerdata}), we implement an inverse square-root frequency reweighing scheme to emphasize the learning of less-frequent classes.  More formally, the loss on a sample is:
\begin{align}
    \mathcal{L}(y,\hat y) &= \frac{1}{C} \sum_{i=0}^{C-1} w_i(y_i) BCE(y_i,\hat y_i) \\
    w_i(z) &= \frac{1}{2} \left(\frac{z}{r_i} + \frac{1-z}{1-r_i}\right) \\
    r_i &= \frac{\sqrt{N_i}}{\sqrt{N_i} + \sqrt{N-N_i}}
\end{align}
where $C$ is the number of classes, $\hat y\in [0,1]^C$ is the prediction, $y\in \{0,1\}^C$ is the ground truth label, $BCE$ is binary cross entropy loss, $N$ is the total number of samples, and $N_i$ is the number of positive samples in the $i^{\text{th}}$ class.  We found that plain inverse frequency weighing caused numerical instability in training, necessitating the square root.

\subsection{Character Segmentation \& Bounding Boxes} \label{sec:bbox}
In order to produce bounding boxes around each subject in the image, we first train an illustrated character segmenter.  As we assume one subject per image, we can derive a bounding box by enclosing the thresholded segmentation output.  The single-subject assumption also removes the need for region proposal and NMS infrastructure present in available illustrated segmenters \cite{aniseg}, so that our model may focus on producing clean segmentations only.  Our segmentation model is based on DeepLabv3 \cite{deeplab}, with three additional layers at the end of the head for finer segmentations at the input image resolution.  We initialize with pretrained DeepLabv3 weights from PyTorch \cite{pytorch}, and fine-tune the full model using pixel-wise binary cross-entropy loss.


\setlength{\tabcolsep}{5pt}
\begin{table*}[!htb]
\begin{center}
\begin{tabular}{|l|r r r r r|r r|}
    \hline
    Model                         & OKS@50  & OKS@75  & PCKh@50 & PDJ@20   & PCPm@50 & params & ms/img \\
    \hline\hline
    Feature Concatenation (+new data) &  \textbf{0.8982} &  \textbf{0.7930} &  \textbf{0.7866} &  0.8403  &  0.8551 & 86.8m & 217.7 \\
    Feature Concatenation         &  0.8827 &  0.7723 &  0.7762 &  0.8282  &  0.8435 & 86.8m & 217.7 \\
    Feature Matching (+new data)  &  0.8953 &  0.7907 &  0.7851 &  \textbf{0.8423}  &  \textbf{0.8599} & 9.9m & 147.8 \\
    Feature Matching              &  0.8769 &  0.7680 &  0.7675 &  0.8251  &  0.8343 & 9.9m & 147.8 \\
    \hline
    Task Fine-tuning Only         &  0.8026 &  0.6481 &  0.7032 &  0.7666  &  0.7446 & 77.5m & 174.5 \\
    Domain Features Only          &  \textbf{0.8607} &  \textbf{0.7467} &  0.7444 &  0.8076  &  \textbf{0.8215} & 9.6m & 143.7 \\
    Task Fine-tuning w/ Domain Features &  0.8548 &  0.7209 &  \textbf{0.7544} &  \textbf{0.8181}  &  0.8084 & 41.1m & 147.8 \\
    Adversarial (DeepFashion2)    &  0.8321 &  0.6804 &  0.7108 &  0.7823  &  0.7778 & 9.9m & 147.8 \\
    Adversarial (COCO)            &  0.8065 &  0.6362 &  0.6788 &  0.7607  &  0.7350 & 9.9m & 147.8 \\
    \hline
    Task-Pretrained (R-CNN)       &  \textbf{0.7584} &  \textbf{0.6724} &  \textbf{0.6960} &  \textbf{0.7357}  &  \textbf{0.6679} & 77.5m & 174.5 \\
    Task-Pretrained (OpenPose)    &  0.4922 &  0.4222 &  0.4447 &  0.4796  &  0.4381 & 52.3m & 128.2 \\
    \hline
    Ours (equiv. to feat. concat.) &  \textbf{0.8827} &  \textbf{0.7723} &  \textbf{0.7762} &  \textbf{0.8282}  &  \textbf{0.8435} & 86.8m & 217.7 \\
    RF5 Backbone                  &  0.8547 &  0.7358 &  0.7427 &  0.8015  &  0.8005 & 86.8m & 217.7 \\
    ImageNet-pretrained Backbone  &  0.8218 &  0.6919 &  0.7060 &  0.7649  &  0.7571 & 86.8m & 217.7 \\
    \hline
\end{tabular}
\end{center}
\caption{Performance of different architectures and ablations described in Sec. \ref{sec:archs}.  Note that the parameter count and speed are measured in inference mode with batch size one; ``m" refers to ``millions of parameters".}
\label{tab:inferperf}
\end{table*}

\section{Data Collection}
Unless mentioned otherwise, we train with random image rotation, translation, scaling, flipping, and recoloring.

\subsection{Pose Data} \label{sec:posedata}
We extend the AnimeDrawingsDataset (ADD), first collected by Khungurn \etal \cite{manpu2016}.  The original dataset had 2000 illustrated full-body single-character images from Danbooru, each annotated with joint keypoints.  However, ADD did not follow the now popularized COCO standard \cite{coco}; in particular, it was missing facial keypoints (eyes and ears) and bounding boxes.  In order to evaluate and compare with modern pose estimators, we manually labeled the missing keypoints using an open-source COCO annotator \cite{cocoannotator} and automatically generated bounding boxes using the character segmenter described in Sec. \ref{sec:bbox}.  We also manually remove 57 images with multiple characters, or without the full body in view.

In addition, we improve the diversity of poses in ADD by collecting an additional 2043 samples.  A major weakness of ADD is its lack of backwards-facing characters; only 5.45\% of the entire 2k dataset had a back-related Danbooru tag (e.g. back, from\_behind, looking\_back, etc.).  We specifically filtered for back-related images when annotating, resulting in a total of 850 in the updated dataset (21.25\%).  We also selected for other notably under-represented poses, like difficult leg tags (soles, bent\_over, leg\_up, crossed\_legs, squatting, kneeling, etc.), arm tags (stretch, arms\_up, hands\_clasped, etc.), and lying tags (on\_side, on\_stomach).

Our final updated dataset contains 4000 illustrated character images with all 17 COCO keypoints and bounding boxes.  We designate 3200 images for training (previously 1373), 313 for validation (previously 97), and 487 for testing (same as original ADD).  For each input image, we first scale and crop such that the bounding box is centered and padded by at least 10\% of the edge length on all sides.  We then perform augmentations; flips require swapping left-right keypoints, and full 360-degree rotations are allowed.

\subsection{ResNet Tagger Data} \label{sec:taggerdata}
Our ResNet50 tagger is trained on a new subset of the 512px SFW Danbooru2019 dataset \cite{danbooru2020}.  The original dataset contains 2.83m images with over 390k tags, but after filtering and retagging we arrive at 837k images with 1062 classes.  The new classes are derived from manually-selected union rules over 2027 raw tags, as described in Sec. \ref{sec:tagger}; the rulebook has 314 body-part, 545 clothing, and 203 miscellaneous (e.g. image composition) classes.  

To combat the class imbalance problem described in Sec. \ref{sec:tagger}, we also rigorously filtered the dataset.  We remove all images that are not single-person (solo, 1girl, or 1boy), are comics (comic, 4koma, doujinshi, etc.), or are smaller than 512px.  Most critically, we remove all images with less than 12 positive tags; these images are very likely under-tagged, and would have introduced many false-negatives to the ground truth.  The final subset of 837k images has significantly reduced class imbalance (median class frequency 0.38\%, minimum 0.04\%) compared to the datasets of available taggers (median 0.07\%, min 0.01\%) \cite{rf5}.

We split the dataset 80-10-10 train-val-test.  As some tags are color-sensitive, we do not jitter the hue; similarly as some tags are orientation-sensitive, we allow up to 15-degree rotations and horizontal flips only.

\subsection{Character Segmentation Data} \label{sec:segdata}
To obtain character bounding boxes, we train a character segmentation model and enclose output regions at 0.5 threshold (Sec. \ref{sec:bbox}).  The inputs to our segmentation system are augmented composites of RGBA foregrounds (with transparent backgrounds) onto RGB backgrounds; the synthetic ground truth is the foreground alpha.  The available AniSeg dataset \cite{aniseg} has only 945 images, with manually-labeled segmentations that are not pixel-perfectly aligned.  We thus collect our own larger synthetic compositing dataset.  Our background images are a mix of illustrated scenery (5.8k Danbooru images with scenery and no\_humans tag) and stock textures (2.3k scraped \cite{pixivutil} from the Pixiv Dataset \cite{pixiv}).  We collect single-character foreground images from Danbooru with the transparent\_background tag; 18.5k samples are used, after filtering images with text, non-transparency, or more than one connected component in the alpha channel.  Counting each foreground as a single sample, this makes our new dataset roughly 20x larger than AniSeg.  The foregrounds and backgrounds are randomly paired for compositing during training, with 5\% chance of having no foreground.  We hold out 2048 deterministic foreground-background pairs for validation and testing (1024 each).

\setlength{\tabcolsep}{3pt}
\begin{table*}[!htb]
\begin{center}
\begin{tabular}{|lr|rrrrrrrrrrrrr|rrrr|}
\hline
keypoint  &&& \multicolumn{2}{c}{OKS@50} && \multicolumn{2}{c}{OKS@75} && \multicolumn{2}{c}{PCKh@50} && \multicolumn{2}{c}{PDJ@20} && \multicolumn{4}{c|}{PDJ@20 \cite{manpu2016}} \\
\hline\hline
nose      &&& 0.9466&(+0.4\%)& & 0.8419&(+3.8\%)& & 0.9918&(+0.2\%)& & 0.9897&(+0.2\%)& &&  0.794&(+24.7\%) & \\
eyes      &&& 0.9795&(+1.1\%)& & 0.9363&(+4.3\%)& & 0.9928&(+0.0\%)& & 0.9928&(+0.1\%)& && *0.890&(+11.6\%) & \\
ears      &&& 0.9589&(+1.3\%)& & 0.8573&(+0.8\%)& & 0.9836&(+0.1\%)& & 0.9795&(-0.2\%)& && *0.890&(+10.1\%) & \\
shoulders &&& 0.9825&(+2.8\%)& & 0.9240&(+1.8\%)& & 0.8973&(+2.6\%)& & 0.9343&(+2.0\%)& && *0.786&(+18.9\%) & \\
elbows    &&& 0.8655&(+3.8\%)& & 0.7320&(+6.4\%)& & 0.7290&(+5.7\%)& & 0.7916&(+4.2\%)& &&  0.641&(+23.5\%) & \\
wrists    &&& 0.7341&(+2.0\%)& & 0.5657&(+2.4\%)& & 0.6263&(+1.2\%)& & 0.6961&(+1.5\%)& &&  0.503&(+38.4\%) & \\
hips      &&& 0.9630&(+0.0\%)& & 0.8686&(+2.8\%)& & 0.6704&(-1.1\%)& & 0.7854&(+0.7\%)& && *0.786&(-0.1\%) & \\
knees     &&& 0.8686&(+2.8\%)& & 0.7444&(+2.5\%)& & 0.6643&(+2.9\%)& & 0.7577&(+3.4\%)& &&  0.610&(+24.2\%) & \\
ankles    &&& 0.8090&(+1.3\%)& & 0.6910&(-0.3\%)& & 0.6263&(+1.0\%)& & 0.7105&(+1.8\%)& &&  0.596&(+19.2\%) & \\
\hline
\end{tabular}
\end{center}
\caption{Keypoint breakdown of our most performant ``feature concatenation" model trained on our extended ADD dataset.  In the center, we list the relative improvement of each metric when training on additional data.  On the right, we display the PDJ@20 from Khungurn \etal \cite{manpu2016}, and report the relative difference from our best model.  *Note that due to keypoint incompatibilities, we fill missing keypoint results from \cite{manpu2016} using the most similar keypoints reported: ``head" for eyes and ears, and ``body" for shoulders and hips.}
\label{tab:inferbody}
\end{table*}

\setlength{\tabcolsep}{6pt}
\begin{table}[]
\centering
\begin{tabular}{|l|r r r r|}
    \hline
    Model                & F-1 & pre. & rec. & IoU \\  
    \hline\hline
    Ours                 & \textbf{0.9472} & \textbf{0.9427} & \textbf{0.9576} & \textbf{0.9326} \\  
    YAAS SOLOv2          & 0.9061 & 0.9003 & 0.9379 & 0.9077 \\
    YAAS CondInst        & 0.8866 & 0.8824 & 0.8999 & 0.9158 \\
    AniSeg               & 0.5857 & 0.5877 & 0.5954 & 0.6651 \\  
    \hline
\end{tabular}
\caption{Comparison of our character segmentation and bounding box performance, described in Sec. \ref{sec:segperf}.}
\label{tab:segperf}
\end{table}

\section{Experiments}

We used PyTorch \cite{pytorch} wrapped in Lightning \cite{lightning}; some models use the R101-FPN keypoint detection R-CNN from Detectron2 \cite{wu2019detectron2}.  All models can be trained with a single GTX1080ti (11GB VRAM).  Unless otherwise mentioned, we trained models using the Adam \cite{adam} optimizer, with 0.001 learning rate and batch size 32, for $1{,}000$ epochs.

The ResNet backbone is trained on the Danbooru tag classification task using our new manual tagging rulebook (Sec. \ref{sec:taggerdata}).  The character segmenter used for bounding boxes is trained with our new character segmentation dataset (Sec. \ref{sec:segdata}).  Using the previous two submodules, we train the pose estimator using our upgraded version of the ADD dataset (Sec. \ref{sec:posedata}).  All data and code will be released upon publication.

\subsection{Pose Estimation Transfer} \label{sec:archs}
Table \ref{tab:inferperf} shows the performance of different architectures.  We report COCO OKS \cite{coco}, PCKh and PCPm \cite{mpipose}, and PDJ (for comparison with Khungurn \etal \cite{manpu2016}).  From the top four rows, we see that our proposed feature concatenation and matching models perform the best out overall, and that the addition of our new data increases performance.  We also observe that while concatenation performs marginally better than matching, matching is 8.8x more parameter efficient and one-third faster at inference.

The second group of Table \ref{tab:inferperf} shows other architectures, roughly in order of method complexity.  Here, as in Fig. \ref{fig:schematic}, ``task" source features refer to Mask R-CNN pose estimation features, and ``domain" source features refer to illustration features extracted by our ResNet50 tag classifier.

\textbf{``Task Fine-tuning Only"} fine-tunes the pretrained Mask R-CNN head with its frozen default backbone; the last head layer is re-initialized to accommodate auxiliary appendage keypoints.  This is vanilla transfer by fine-tuning a task-specific source network on a small task-specific target domain dataset.

\textbf{``Domain Features Only"} is our frozen ResNet50 backbone with a keypoint head.  This is vanilla transfer by adding a new task head to a domain-specific source network.

\textbf{``Task Fine-tuning w/ Domain Features"} fine-tunes the pretrained Mask R-CNN head as above, but replaces the R-CNN backbone with our frozen ResNet50 backbone.  This is a naive method of incorporating both sources, attempting to adapt the task source's pretrained prediction component to new domain features.

\textbf{``Adversarial (DeepFashion2)"} reuses the feature matching architecture, but performs adversarial domain discrimination instead of MSE matching.  The discriminator is a shallow 2-layer convnet, trained to separate Mask R-CNN features of randomly sampled DeepFashion2 \cite{deepfashion2} images from ResNet features of Danbooru illustrations.  As the feature maps to discriminate are spatial, we are careful to employ only 1x1 kernels in the discriminator; otherwise, the discriminator could pick up intrinsic anatomical differences.  The matching network now fools the discriminator by adversarially aligning the feature distributions.

\textbf{``Adversarial (COCO)"} is the same adversarial architecture as above, but using COCO \cite{coco} images containing people instead of Deepfashion2.

\begin{figure*}[!htb]
\begin{center}
    \includegraphics[width=\linewidth]{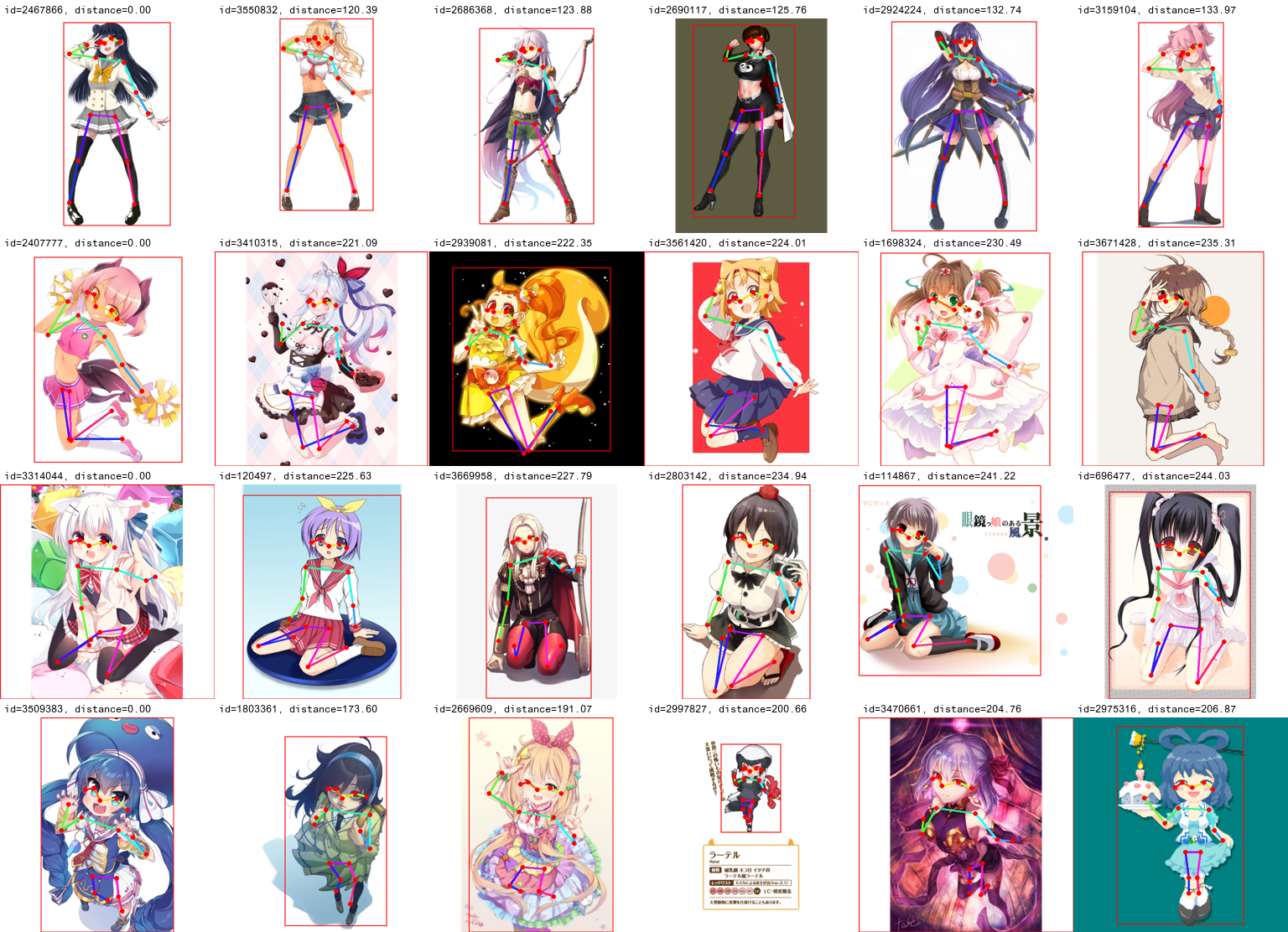}
    \end{center}
    \caption{Pose-based retrieval.  From left to right, we show the query image (descriptor distance zero) followed by its five nearest neighbors (duplicate and NSFW images removed).  Each illustration is annotated with its Danbooru ID, descriptor distance to the query, and the predicted bounding box with COCO keypoints.  Please see supplementary materials for full artist attribution and additional examples.}
\label{fig:retrieve}
\end{figure*}

While domain-features-only is the cheapest architecture overall, it is only slightly more efficient than feature matching, and loses all benefits of task-specific transfer.  However, the performance drop from feature concatenation to domain-features-only and task-with-domain-features is not very large (2-3\% OKS@50); meanwhile, there is a wide gap to task-fine-tuning-only.  This shows that the domain-specific ResNet50 backbone trained on our new body-tag rulebook provides much more predictive power than the task-specific pretrained Mask R-CNN.

It is important to note that the adversarial models exhibited significant instability during training.  After extensive hyperparameter tuning, the best DeepFashion2 model returns NaN loss at epoch 795, and the best COCO model fails at epoch 354; all other models safely exited at epoch $1{,}000$.  DeepFashion2 likely outperforms COCO because the image composition is much more similar to that of Danbooru; images are typically single-person portraits with most of the body in view.  Adversarial losses are notoriously difficult to optimize, and in our case destabilized training so as to perform worse than not having been used at all.

The fourth group of Table \ref{tab:inferperf} shows out-of-the-box generalization to illustrations for Mask R-CNN \cite{rcnn} and OpenPose \cite{paf}.  We use Mask R-CNN as our task-specific source, as it is less-overfit to natural images than OpenPose.

Table \ref{tab:inferbody} gives a keypoint breakdown and comparison with Khungurn \etal \cite{manpu2016}.  The results demonstrate that training on our additional more varied data improves the overall model performance; this is especially true for appendage keypoints, which are more variable than the head and torso.  We also see significant improvement from results reported in Khungurn \etal.  The exception is the hips, for which we compare to their ``body" keypoint at the navel.  While this is not a direct comparison, our PDJ on hips is nevertheless low relative to other keypoints.  This is because PDJ does not account for the intrinsic ambiguity of the hips; looking at the OKS, which accounts for annotator disagreement, we see that hip performance is actually quite high.

An important caveat is that the metrics are generally not comparable with those reported in human pose estimation.  COCO OKS, for example, was designed using annotator disagreement on natural images \cite{coco}; however, illustrated character proportions deviate widely from the standard human form (i.e. bigger head and eyes).  Characters also tend to take up more screen space proportional to body size (i.e. big hair and clothing), leading to looser thresholds normalized by bounding box size.

\subsection{ResNet Tagger Backbone}
We train our ResNet50 tagger backbone to produce illustration-specific source features (Fig. \ref{fig:schematic}).  Taking into account the class imbalance, we accumulate gradients for an effective batch size of 512.  Considering the minimum (0.04\%) and median (0.38\%) class frequencies, we may expect the smallest class to appear 0.2 times per batch, and the median class to appear 1.9 times per batch.

To demonstrate the effectiveness of our tag rulebook and class reweighing strategy, we report performance on pose estimation using two other ResNet50 backbones: the RF5 tagger \cite{rf5}, and the default ImageNet-pretrained ResNet50 from PyTorch \cite{pytorch}.  While there are several Danbooru taggers available \cite{rf5, deepdanbooru}, we chose to compare our backbone to the RF5 tagger \cite{rf5} because it is the most architecturally similar to our ResNet50, and relatively better-documented.  The backbones all share the same architecture and parameter count, and are all placed into our feature concatenation transfer model for the ablation.

The backbone ablation results are shown in the last three rows of Table \ref{tab:inferperf}.  As expected, a classifier trained with our novel body-part-specific tagging rulebook and class-balancing techniques significantly improves transfer to pose estimation. Note that our tagger also outperforms RF5 at classification (on shared target classes); please refer to the supplementary materials for more details.

\subsection{Character Segmentation \& Bounding Boxes} \label{sec:segperf}
We compare the segmentation and bounding box performance of our system with that of publicly-available models.  AniSeg \cite{aniseg} is a Faster-RCNN \cite{frcnn}, and YAAS \cite{yaas} provides SOLOv2 \cite{solo2} and CondInst \cite{condinst} models.  These detectors may detect more than one character, and their bounding boxes are not necessarily tight around segmentations; for simplicity, we union all predicted segmentations of an image, and redraw a tight bounding box around the union.  We evaluate all models on the same test set described in Sec. \ref{sec:segdata}.  Table \ref{tab:segperf} shows that training with our new 20x larger dataset outperforms available models in both mean F-1 (segmentation) and IoU (bounding boxes); we thus use it in our pipeline for bounding box prediction.

\section{Application: Pose-guided Retrieval}

An immediate application of our illustrated pose estimator is a pose-guided character retrieval system.  We construct a proof-of-concept retriever that takes a query character (or user-specified keypoints and bounding box) and searches for illustrated characters in a similar pose.  This system can serve as a useful search tool for artists, who often use reference drawings while illustrating.

Our pose retriever performs a simple nearest-neighbor search.  The support images consist of single-character Danbooru illustrations with the full\_body tag.  Using our best-performing model, we extract bounding boxes and keypoint locations for each character, normalize the keypoints by the longest bounding box dimension, and finally store the pairwise euclidean distances between the normalized keypoints.  This process ensures the pairwise-distance descriptor is invariant to translation, rotation, and image scale.  At inference, we extract the descriptor from the query, and find the euclidean k-nearest neighbors from the support set.

In practice, we compute descriptors using all 25 predicted keypoints (17 COCO and 8 additional appendage midpoints).  This makes the descriptor 300-dimensional (25 choose 2), which is generally too large for tree-based nearest neighbors \cite{sklearn_api}.  However, since our support set consists of 136k points, we are still able to brute force search in reasonable time.  Empirically, each query takes about 0.1341s for keypoint extraction (GPU) and 0.0638s for search (CPU).

To demonstrate the effectiveness of our pose estimator, we present several query results in Fig. \ref{fig:retrieve}; while there is no ground-truth to measure quantitative performance, qualitative inspection suggests that our model works well.  We can retrieve reasonably similar illustrations for standard poses as shown in the first row, as well as more difficult poses for which illustrators would want references.  Note that while our system has no awareness of perspective, it is able to effectively leverage keypoint cues to retrieve similarly foreshortened views in the last row.  For more examples, please refer to our supplementary materials.

\section{Conclusion \& Future Work}

While we may continue to improve the transfer performance through methods like pseudo-labeling \cite{animals} or cycle-consistent image translation \cite{cycada}, we can also begin extending our work to multi-character detection and pose estimation.  While it is possible to construct a naive instance-based segmentation and keypoint estimation dataset by compositing background-removed ADD samples, we cannot expect a system trained on such data to perform well in-the-wild.  Character interactions in illustrations are often much more complex than human interactions in real life, with much more frequent physical contact.  For example, Danbooru has 43.6k images tagged with holding\_hands and 59.1k with hugging, already accounting for 2.8\% of the entire dataset.  Simply compositing independent characters together would not be able to model the intricacies of the illustration domain; we would again need to expand our datasets with annotated instances of character interactions.

As a fundamental vision task, pose estimation also provides a valuable prior for numerous other novel applications in the illustrated domain.  Our pose estimator opens the door to pose-guided retargeting for automatic character animation, better keyframe interpolation, pose-aware illustration colorization, 3D character reconstruction, etc.

In conclusion, we demonstrate state-of-the-art pose estimation on the illustrated character domain, by leveraging both domain-specific and task-specific source models.  Our model significantly outperforms prior art \cite{manpu2016} despite the absence of synthetic supervision, thanks to successful transfer from our new illustration tagging subtask focused on classifying body-related tags.  In addition, we provide a single-region proposer trained on a novel character segmentation dataset 20x larger than those currently available, as well as an updated illustration pose estimation dataset with twice the number of samples in more diverse poses.  Our model performance allows for the novel task of pose-guided character illustration retrieval, and paves the way for future applications in the illustrated domain.


{\small
\bibliographystyle{ieee_fullname}
\bibliography{egbib}
}

\raggedbottom
\onecolumn


\section{Supplementary Materials}

\subsection{Tagger Classification Comparison}
In the main paper, we show that our tagger (trained on our new tag rulebook with class-balanced weighing) significantly improves transfer to pose estimation.  Here, we show classification results in comparison to the RF5 Danbooru tagger \cite{rf5}, a publicly-available model with the same ResNet50 architecture.  RF5 predicts the presence of the top 6000 most common tags in the dataset; 1207 of these are present in our new rulebook, and can be used to predict 1032 of the 1062 total new classes.  As we can see from Table \ref{tab:tagperf} below, our model performs much better at classifying the same tags.

\setlength{\tabcolsep}{10pt}
\begin{table*}[!htb]
\centering
\begin{tabular}{|l|r r|}
    \hline
    Model      & Ours & RF5 \\
    \hline\hline
    F-2        & 0.4744 & 0.2297 \\
    precision  & 0.3022 & 0.1238 \\
    recall     & 0.5786 & 0.3360 \\
    accuracy   & 0.9760 & 0.9496 \\
    \hline
    F-1        & 0.4249 & 0.1910 \\
    precision  & 0.4236 & 0.1898 \\
    recall     & 0.4458 & 0.2235 \\
    accuracy   & 0.9851 & 0.9727 \\
    \hline
\end{tabular}
\caption{Comparison of our Danbooru tagger to RF5 \cite{rf5}.  Metrics are calculated using per-class optimal thresholds for either F-1 or F-2, and averaged across all classes shared between models.  Note that this means F-1 and F-2 cannot be directly calculated from their respective precision and recall statistics in the table.}
\label{tab:tagperf}
\end{table*}

\subsection{Pose Retrieval Additional Results}
\begin{figure*}[!ptb]
\begin{center}
    \includegraphics[width=\linewidth]{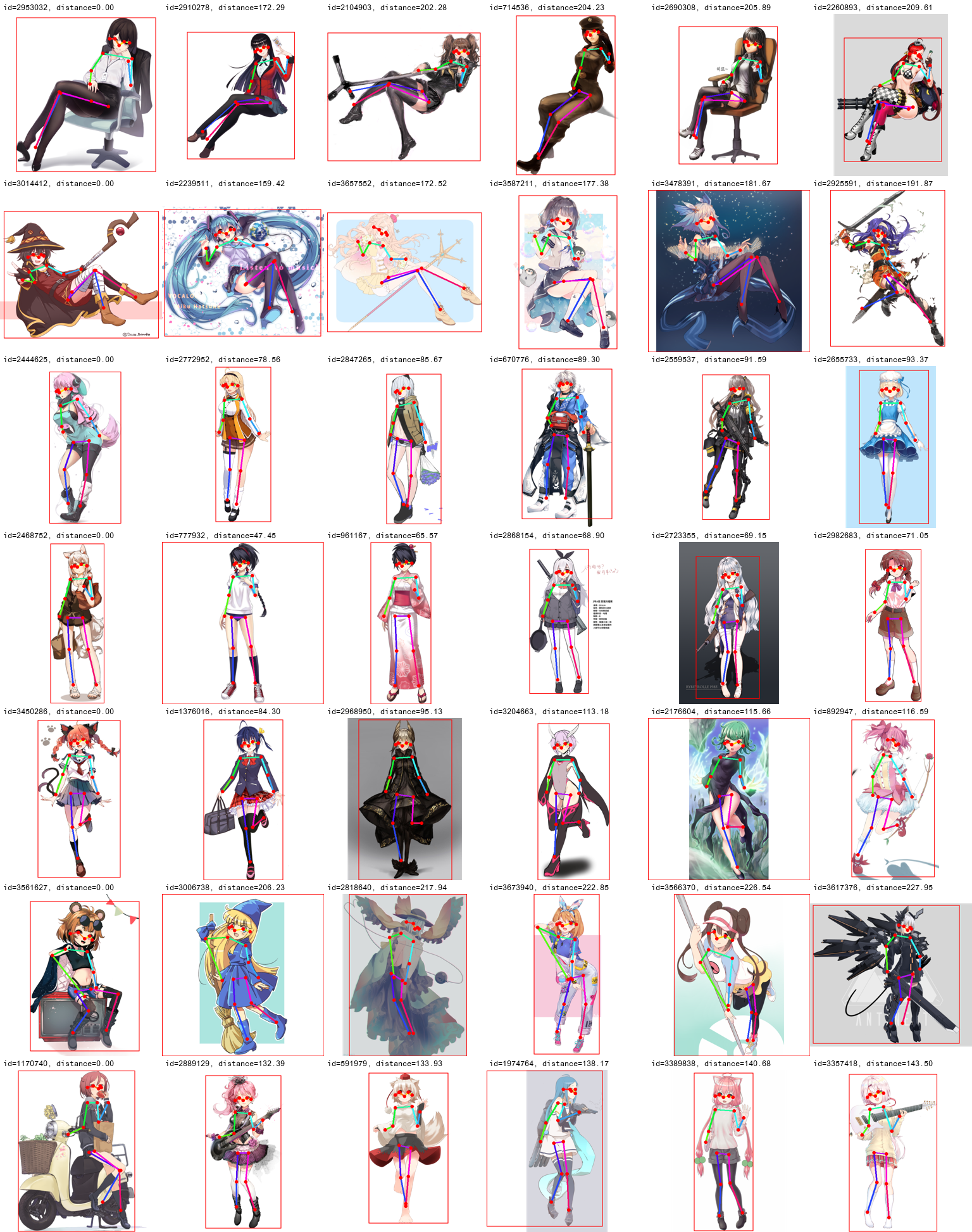}
    \end{center}
    \caption{Additional pose-based retrieval results.  From left to right, we show the query image (descriptor distance zero) followed by its five nearest neighbors (duplicate and NSFW images removed).  Each illustration is annotated with its Danbooru ID, descriptor distance to the query, and the predicted bounding box with COCO keypoints.}
\label{fig:retrievesupp}
\end{figure*}

We display several more pose-based illustration retrieval results in Fig. \ref{fig:retrievesupp}; the images are taken from the Danbooru dataset \cite{danbooru2020}.  The first two rows show challenging sitting positions, on which our model still performs well qualitatively.  Despite the differences in orientation, our rotation-invariant descriptor is still able to identify the poses as similar.  Rows 3-5 show some more standard poses.  Notice that in row 4, the first and second neighbors are variations of the same character in the same pose; it is very common to find a set of such variations uploaded to Danbooru together, and our model may help identify them.  In the last two rows, we show failure cases of our model, where incorrect predictions on the query result in neighbors with different poses.

We also provide artist attributions for the figures in the main paper and this supplementary document.  Danbooru URLs are at \texttt{https://danbooru.donmai.us/posts/DANBOORU\_ID}.  Artist websites are tracked from the illustration's Danbooru page to the best of our ability.  Please note the links might not be SFW.

\setlength{\tabcolsep}{5pt}
\begin{table*}[!htb]
\begin{center}
\begin{tabular}{|r r | l l|}
    \hline
    Row,Col & Danbooru ID & Artist Handle & Artist URL \\
    \hline\hline
0,0 & 2467866 & \begin{CJK}{UTF8}{min} 露茶 \end{CJK} & https://www.pixiv.net/en/users/263905 \\
0,1 & 3550832 & \begin{CJK}{UTF8}{min} Goｰ1 \end{CJK} & https://twitter.com/go\_1tk \\
0,2 & 2686368 & \begin{CJK}{UTF8}{min} あれっくす \end{CJK} & https://twitter.com/alexmaster55 \\
0,3 & 2690117 & \begin{CJK}{UTF8}{min} クラモリ \end{CJK} & https://www.pixiv.net/en/users/1209275 \\
0,4 & 2924224 & \begin{CJK}{UTF8}{min} 紅葉(くれは)＠お仕事募集中 \end{CJK} & https://www.pixiv.net/en/users/6006540 \\
0,5 & 3159104 & \begin{CJK}{UTF8}{min} みの字 \end{CJK} & https://www.pixiv.net/en/users/1523486 \\
1,0 & 2407777 & \begin{CJK}{UTF8}{min} 唯川 \end{CJK} & https://twitter.com/yUikw \\
1,1 & 3410315 & \begin{CJK}{UTF8}{min} 楠本みや \end{CJK} & https://www.pixiv.net/en/users/6211628 \\
1,2 & 2939081 & \begin{CJK}{UTF8}{min} SeNMU \end{CJK} & https://www.pixiv.net/en/users/65308 \\
1,3 & 3561420 & \begin{CJK}{UTF8}{min} なもり@ゆるゆり20巻でた \end{CJK} & https://twitter.com/\_namori\_ \\
1,4 & 1698324 & \begin{CJK}{UTF8}{min} ratryu \end{CJK} & https://www.pixiv.net/en/users/3892817 \\
1,5 & 3671428 & \begin{CJK}{UTF8}{min} ぼや野 \end{CJK} & https://www.pixiv.net/en/users/1263092 \\
2,0 & 3314044 & \begin{CJK}{UTF8}{min} 九条だんぼ \end{CJK} & https://twitter.com/\_Dan\_ball \\
2,1 & 120497 & \begin{CJK}{UTF8}{min} 劉祥 \end{CJK} & https://www.pixiv.net/en/users/22017 \\
2,2 & 3669958 & \begin{CJK}{UTF8}{min} もやし \end{CJK} & https://twitter.com/moyashi\_mou2 \\
2,3 & 2803142 & \begin{CJK}{UTF8}{min} エノキドォ \end{CJK} & https://www.pixiv.net/en/users/4535430 \\
2,4 & 114867 & \begin{CJK}{UTF8}{min} なまもななせ・海通信 \end{CJK} & https://www.pixiv.net/en/users/1167548 \\
2,5 & 696477 & \begin{CJK}{UTF8}{min} みぞれまじ \end{CJK} & https://www.pixiv.net/en/users/1502612 \\
3,0 & 3509383 & \begin{CJK}{UTF8}{min} 和菓子 \end{CJK} & https://www.pixiv.net/en/users/13748172 \\
3,1 & 1803361 & \begin{CJK}{UTF8}{min} アチャコ \end{CJK} & https://www.pixiv.net/en/users/1302618 \\
3,2 & 2669609 & \begin{CJK}{UTF8}{min} Seedkeng \end{CJK} & https://www.pixiv.net/en/users/11039166 \\
3,3 & 2997827 & \begin{CJK}{UTF8}{min} 吉崎 観音 \end{CJK} & https://twitter.com/yosRRX \\
3,4 & 3470661 & \begin{CJK}{UTF8}{min} テイク \end{CJK} & https://www.pixiv.net/en/users/2096681 \\
3,5 & 2975316 & \begin{CJK}{UTF8}{min} いたたたた \end{CJK} & https://twitter.com/itatatata6 \\

    \hline
\end{tabular}
\end{center}
\caption{Artist attribution for Figure \ref{fig:retrieve} of the main paper.}
\label{tab:artists_main}
\end{table*}

\setlength{\tabcolsep}{5pt}
\begin{table*}[!htb]
\begin{center}
\begin{tabular}{|r r | l l|}
    \hline
    Row,Col & Danbooru ID & Artist Handle & Artist URL \\
    \hline\hline
0,0 & 2953032 & \begin{CJK}{UTF8}{min} Dev@プロフィールを確認!! \end{CJK} & https://www.pixiv.net/en/users/857300 \\
0,1 & 2910278 & \begin{CJK}{UTF8}{min} shrimpえびちゃん \end{CJK} & https://www.pixiv.net/en/users/3989209 \\
0,2 & 2104903 & \begin{CJK}{UTF8}{min} 枚方ガルダイン \end{CJK} & https://twitter.com/hgd\_AG \\
0,3 & 714536 & \begin{CJK}{UTF8}{min} steward B \end{CJK} & https://www.pixiv.net/en/users/2212889 \\
0,4 & 2690308 & \begin{CJK}{UTF8}{min} aken@お仕事募集中 \end{CJK} & https://www.pixiv.net/en/users/196317 \\
0,5 & 2260893 & \begin{CJK}{UTF8}{min} ABBB \end{CJK} & https://www.pixiv.net/en/users/4066325 \\
1,0 & 3014412 & \begin{CJK}{UTF8}{min} Disco \end{CJK} & https://www.pixiv.net/en/users/14377769 \\
1,1 & 2239511 & \begin{CJK}{UTF8}{min} 千里GAN（改名） \end{CJK} & https://twitter.com/oshirase\_gan \\
1,2 & 3657552 & \begin{CJK}{UTF8}{min} 塩soda \end{CJK} & https://twitter.com/kurau3 \\
1,3 & 3587211 & \begin{CJK}{UTF8}{min} せんちゃ \end{CJK} & https://www.pixiv.net/en/users/3388329 \\
1,4 & 3478391 & \begin{CJK}{UTF8}{min} ぺろんちょ \end{CJK} & https://www.pixiv.net/en/users/2689378 \\
1,5 & 2925591 & \begin{CJK}{UTF8}{min} 米山舞 \end{CJK} & https://www.pixiv.net/en/users/1554775 \\
2,0 & 2444625 & \begin{CJK}{UTF8}{min} Pack \end{CJK} & https://www.pixiv.net/en/users/6069171 \\
2,1 & 2772952 & \begin{CJK}{UTF8}{min} しずまよしのり \end{CJK} & https://twitter.com/M\_ars \\
2,2 & 2847265 & \begin{CJK}{UTF8}{min} 千羽ミハト(Mihato) \end{CJK} & https://www.pixiv.net/en/users/485524 \\
2,3 & 670776 & \begin{CJK}{UTF8}{min} 芹ざわ\_お仕事募集中 \end{CJK} & https://www.pixiv.net/en/users/1156416 \\
2,4 & 2559537 & \begin{CJK}{UTF8}{min} infukun \end{CJK} & https://www.pixiv.net/en/users/5065896 \\
2,5 & 2655733 & \begin{CJK}{UTF8}{min} QBASE \end{CJK} & https://www.pixiv.net/en/users/246248 \\
3,0 & 2468752 & \begin{CJK}{UTF8}{min} アモニット \end{CJK} & https://www.pixiv.net/en/users/1621147 \\
3,1 & 777932 & \begin{CJK}{UTF8}{min} ヒエラポリスとパムッカレ \end{CJK} & https://www.pixiv.net/en/users/873649 \\
3,2 & 961167 & \begin{CJK}{UTF8}{min} ヒエラポリスとパムッカレ \end{CJK} & https://www.pixiv.net/en/users/873649 \\
3,3 & 2868154 & \begin{CJK}{UTF8}{min} 菟\_ \end{CJK} & https://www.pixiv.net/en/users/3621731 \\
3,4 & 2723355 & \begin{CJK}{UTF8}{min} 9Ro \end{CJK} & https://twitter.com/RINGDDINGDONGX2 \\
3,5 & 2982683 & \begin{CJK}{UTF8}{min} dairi \end{CJK} & https://www.pixiv.net/en/users/4920496 \\
4,0 & 3450286 & \begin{CJK}{UTF8}{min} さとうぽて \end{CJK} & https://twitter.com/mrcosmoov \\
4,1 & 1376016 & \begin{CJK}{UTF8}{min} トレス \end{CJK} & https://www.pixiv.net/en/users/2056112 \\
4,2 & 2968950 & \begin{CJK}{UTF8}{min} LM7 \end{CJK} & https://www.pixiv.net/en/users/420928 \\
4,3 & 3204663 & \begin{CJK}{UTF8}{min} Cowfee@home \end{CJK} & https://twitter.com/cowfee\_desu \\
4,4 & 2176604 & \begin{CJK}{UTF8}{min} ゴールデン＠お仕事募集中 \end{CJK} & https://www.pixiv.net/en/users/8607687 \\
4,5 & 892947 & \begin{CJK}{UTF8}{min} mizuki \end{CJK} & https://www.pixiv.net/en/users/1050881 \\
5,0 & 3561627 & \begin{CJK}{UTF8}{min} keti \end{CJK} & https://www.pixiv.net/en/users/4705322 \\
5,1 & 3006738 & \begin{CJK}{UTF8}{min} ささ吉 \end{CJK} & https://www.pixiv.net/en/users/7187584 \\
5,2 & 2818640 & \begin{CJK}{UTF8}{min} しろもる \end{CJK} & https://www.pixiv.net/en/users/19858643 \\
5,3 & 3673940 & \begin{CJK}{UTF8}{min} Jiujiu girl \end{CJK} & https://www.pixiv.net/en/users/1221354 \\
5,4 & 3566370 & \begin{CJK}{UTF8}{min} ２番目のむ～みん \end{CJK} & https://www.pixiv.net/en/users/8186490 \\
5,5 & 3617376 & \begin{CJK}{UTF8}{min} まかだみぁ \end{CJK} & https://www.pixiv.net/en/users/2782928 \\
6,0 & 1170740 & \begin{CJK}{UTF8}{min} 黒星紅白 \end{CJK} & https://www.pixiv.net/en/users/178217 \\
6,1 & 2889129 & \begin{CJK}{UTF8}{min} Anmi@画集発売中 \end{CJK} & https://www.pixiv.net/en/users/212801 \\
6,2 & 591979 & \begin{CJK}{UTF8}{min} wk. \end{CJK} & https://www.pixiv.net/en/users/142914 \\
6,3 & 1974764 & \begin{CJK}{UTF8}{min} katsuragi (osaka8o2) \end{CJK} & https://danbooru.donmai.us/artists/108387 \\
6,4 & 3389838 & \begin{CJK}{UTF8}{min} るっぴ \end{CJK} & https://twitter.com/gouya\_yu0508 \\
6,5 & 3357418 & \begin{CJK}{UTF8}{min} guitaro (yabasaki taro) \end{CJK} & https://danbooru.donmai.us/artists/181484 \\

    \hline
\end{tabular}
\end{center}
\caption{Artist attribution for Figure \ref{fig:retrievesupp} in the supplementary materials.}
\label{tab:artists_supp}
\end{table*}

\end{document}